# VERIFYING HEAPS' LAW USING GOOGLE BOOKS NGRAM DATA

**Vladimir V. Bochkarev, Eduard Yu.Lerner, Anna V. Shevlyakova**

Kazan Federal University, Kremlevskaya str.18, Kazan 420018, Russia

E-mail: vladimir.bochkarev@kpfu.ru

**Abstract**. This article is devoted to the verification of the empirical Heaps law in European languages using Google Books Ngram corpus data. The connection between word distribution frequency and expected dependence of individual word number on text size is analysed in terms of a simple probability model of text generation. It is shown that the Heaps exponent varies significantly within characteristic time intervals of 60-100 years.

## 1. Introduction

The dependence of lexicon size on a given text size has been studied and widely discussed among linguists and other scholars in the 20th century and is still topical. The most important results obtained in this area relate to Zipf's and Heaps' laws. Heaps' law describes the dependence of individual word number in a text on a given text size and says, in its original formulation, that the number of these words increases as a square root of a number of words in a text. In this paper, we verify Heaps' law in regard to large diachronic text corpora and study the connection between Heaps' and Zipf's laws by performing statistical analyses of a large text corpus from the Google Books Ngram project.

According to Zipf's law, word usage frequency is defined as a power function $p_r \sim r^{-\beta}$, where $r$ is the word rank, in other words the number of the word in a numbered list of words ranked according to their frequency [1]. Zipf's law is closely connected to Heaps' law [2], which says that the size of lexicon $N$ (the number of particular words in a text) in a text or set of texts of size $L$ is determined by $N \sim L^{-k}$. Different probability models of text generation (under the assumption that Zipf's law is fulfilled) result in a simple relation between the exponents β and $k$:

$$k = \beta^{-1} \tag{1}$$

Initially, Zipf's law was tested on a relatively small corpus of texts where the values of exponent $\beta$ were close to 1. Heaps' law was initially derived from the analysis of news items. At that, the exponent $k$ was estimated to be close to 0.5 [3].

Further surveys suggested different generalisations of these laws, including a general case of power dependence. It should also be noted that Heaps' law was formulated (and verified) using text corpora which created during a relatively short period of time. When Heaps' law is applied to large diachronic text corpora additional considerations are necessary.

The release of the large Google Books Ngram corpus, which contains frequency data on 8 languages over a large period of time offered an opportunity to study the static dependencies of words usage [4, 5, 6]. The corpus size is great. For example, it contains $2.94 \cdot 10^6$ English books for a period 1900-2008 which in turn include about $2.39 \cdot 10^{11}$ words. As for the Russian, German and French languages the corpora are also very large. Thus, for these languages there are respectively $3.34 \cdot 10^5$, $3.49 \cdot 10^5$, and $2.59·10^5$ books and $2.58·10^{10}$, $2.50·10^{10}$, and $2.26·10^{10}$ words. Zipf's law is tested in detail on the texts written in the main European languages in [6, 7, 8]. It is shown here that Zipf's law in its classical form is not fulfilled. Rather, a combination of two different power law distributions in different segments of the frequency distribution plot is found. The exponent of the first section (for frequently used words) is close to 1 and the exponent of the second section (for non-frequent words) is much higher and varies from 1.7 to 2.5 for different languages. The last values correlate better with the power values of Heaps' law and expression (1).

This does not mean that the problem of selecting the most suitable and simple model of word usage frequencies is solved. As shown in [6], although the model combining two different power laws fits the empirical data better than the previous models, it has to be rejected for any reasonably selected significant level.

A typical word frequency distribution is shown in Figure 1. The figure shows the word frequency distribution for English (both British and American) words in year 2000. The approximation of the empirical data using power law dependence for the rank ranges 3-440 and $1·10^4$-$5·10^5$ is also shown in the figure. The fitting was done using maximum likelihood estimation under the assumption that the frequency vector is governed by a multinomial distribution. It can be seen that at some segments the power law is not strictly fulfilled. The dependence is rather complicated and is not likely to be adequately described by a simple model with a small number of parameters. It should be noted that the curve goes up in the area of 500-$2·10^4$ ranks. Consequently, when changing the limits used to perform fitting, the values of the power exponent obtained will show a rather large range of variation. The frequencies in the range 3-300 coincide well with a power law. The power exponent, however, is significantly higher than 1 (the exponent is 1.0766 for the case shown in the figure). Using alphabetic and syllabic word-building Markov models, values which are strictly larger than 1 can be obtained [9, 10, 11]. The model coincides best with the empirical data in the low rank area which is rather surprising because it is normal to expect that probability models describe rare words and neologisms better than them.

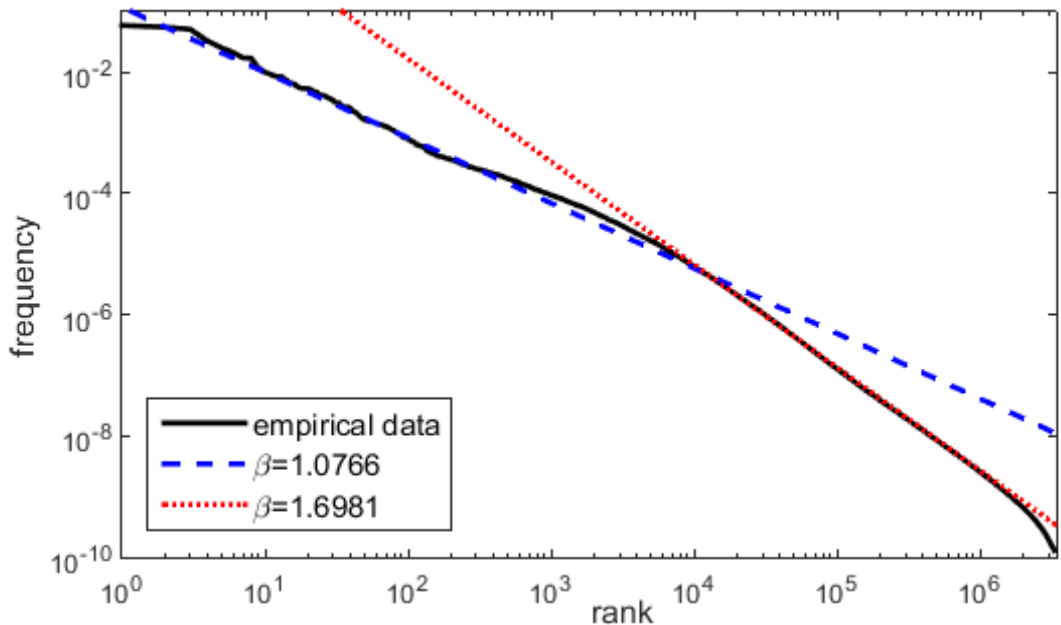

**Figure 1.** Word frequency distribution for English words, year 2000. The solid line represents the empirical data, the dashed and dotted lines are power law approximations.

The frequency distribution within for large ranks is very important for the $N(L)$ dependency explanation. It should be noted that the power law for frequencies from this area cannot be regarded as reliable. The available empirical data do not contradict the assumption about a sub-exponential law which can also be proved using the Markov word-building models [10, 11].

In [6], a model is used to explain the empirical data which assumes an increasing probability of repeated words and a division of the lexicon into a core and a periphery. Let us consider a simpler model of text generation. We assume the existence of a (finite or infinite) set of possible words whose usage probabilities at a subsequent stage is a-priori known and equal to $p_i$. How many words will be used on average at least once in a text of length $L$? To answer this question we will use the method of indicators. Consider a random variable which takes the value 1 if the $i^{th}$ word from our lexicon is used at least once and value 0 if it is not used at all. The probability that the $i^{th}$ word will not be used at least once is $(1-p_i)^L$, consequently our random variable takes value 1 with the probability $1-(1-p_i)^L$ and its mathematical expectation is equal to this number. The mathematical expectation of such random variables is the average number of $N$ words. This yields the following:

$$N(L) = \sum_i \left\{ 1 - (1-p_i)^L \right\} \qquad (2)$$

Thus, we can assess the expected dependence $N(L)$ at the given word frequencies using the model described above.

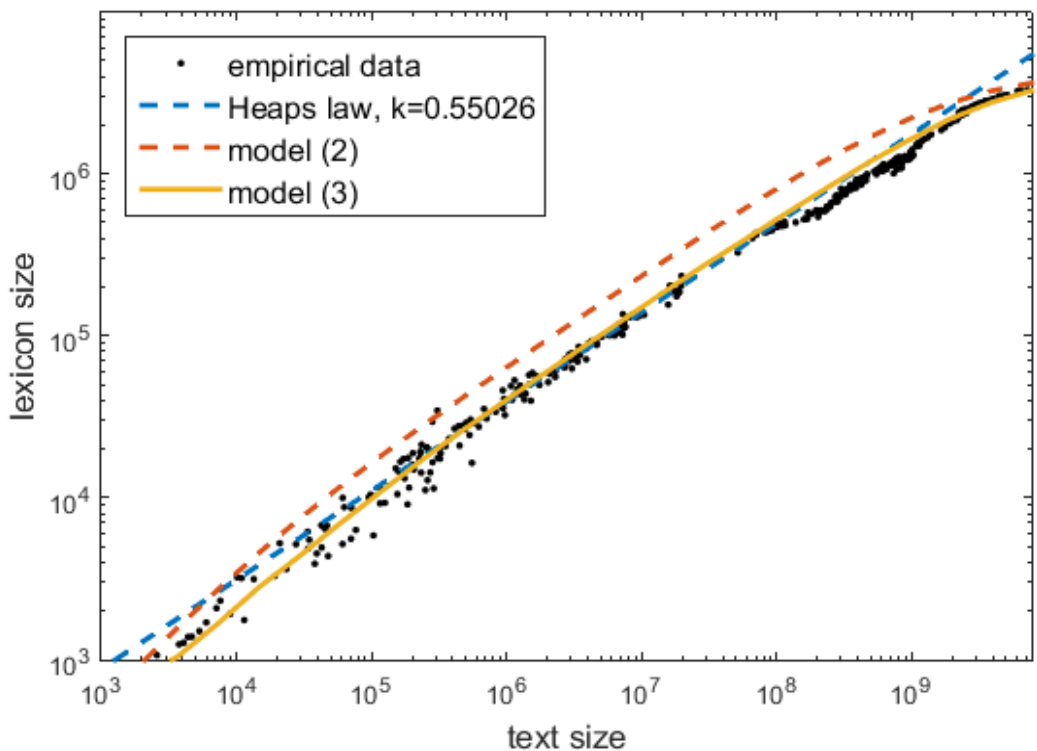

**Figure 2.** The dependence of the lexicon size on the text size in the English data. The dots represent the empirical data and the dotted line the power law approximation; The dashed and solid lines represent the calculation according to the models (2) and (3)

## 2. Comparison of modelled and empirical data

Figure 2 shows the dependence of the lexicon size on the text size according to the Google Books Ngram corpus data for English (British and American combined). The number of different word forms used in the given year (ignoring capitalization) and the total number of words contained in the text base were calculated. Only word forms consisting of the letters of the Latin alphabet were considered. An approximation of the empirical data using power law dependency is also shown in the figure. The power exponent calculated using the least squares method was 0.5503. As it can be seen, the empirical data is poorly described by the exponential function. It should be noted that the correlation with the power law for the other languages presented in the Google Books Ngrams are even worse than for English.

The modelling of the expected lexicon size was carried out using formula (2). It is necessary to know the word-usage frequencies to calculate the expected lexicon size using formula (2). The calculation of frequencies of rare words is relatively problematic especially regarding words used in earlier time periods because the number of texts in Google Books Ngram where they can be found is rather small. Empirical frequencies of words used in the year 2000 were analysed to estimate probabilities $p_k$ because the biggest number of texts is dated back to this year. Thus, we can estimate the frequencies of the greatest number ($3.97 \cdot 10^6$) of unique words. The counting results according to formula (2) are represented in the figure by the dashed line. It can be seen that the modelled dependence is close to the empirical one but it lies slightly above it. To enable a better match of the modelled and experimental data, the model was improved.

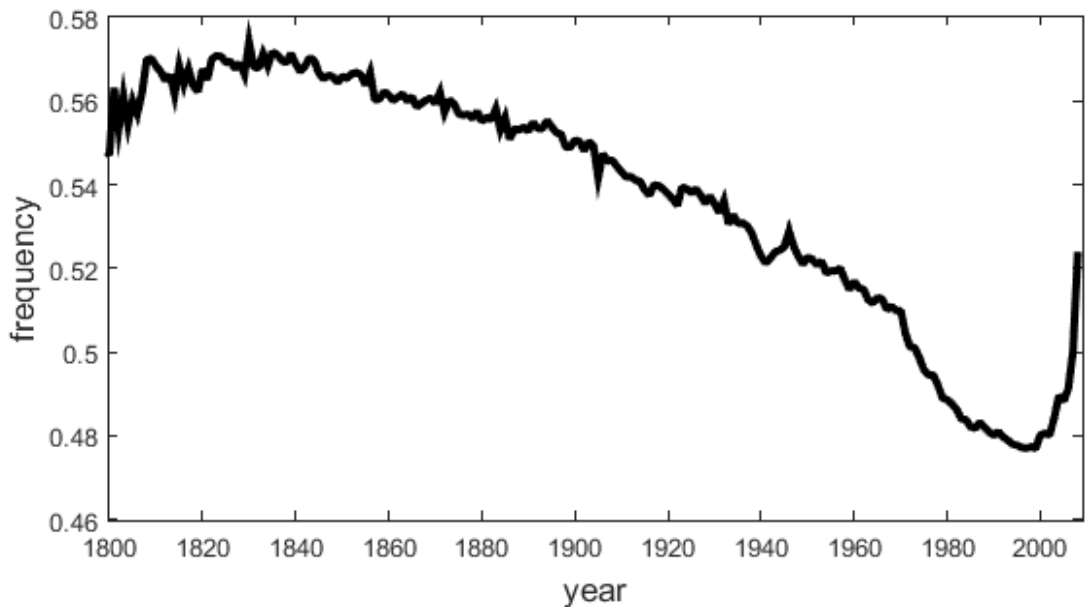

**Figure 3.** Change of functional words percentage in the English language

Traditionally the lexicon is divided into content and function words. The latter indicate a grammatical relationship and their frequency directly depend on the sentence structure. The share of function words in the Google Books Ngram corpus for English is shown in Figure 3 (the list of words described in [12] were used). As it can be seen in the figure, the percentage of function words can be relatively high. For example, the percentage of function words in the English language changed from about 0.48 to 0.57 in 1800-2000. It can be assumed, that all the function words will be used in a set of texts covering the wide range of topics and the number of content words will grow up with the increase of the text size. This results in the following modified model:

$$N(L) = N_{serv} + \sum_{i \in I} \left\{ 1 - (1 - p_i)^{\zeta L} \right\}, \qquad (3)$$

where $N_{serv}$ is the number of function words, $I$ is the set of content word numbers in a common list, and $\zeta$ is the part of content words in a text. In the modelling process, the parameter $\zeta$ is defined for each year according to the Google Books Ngram data, after that the expected lexicon size is calculated using formula (3). The modelling results are shown by the solid line in Figure 2.

It can be seen that the model (3) yields the best approximation of the empirical data. The exponent obtained during the fitting procedure of the modelled curve using the power dependence (on the segment from $10^3$ to $10^{10}$) is 0.5674. This is close to the value mentioned above obtained during the fitting of the empirical data. At the same time, a serious divergence of the modelled and empirical curves can be seen. For example, variations in the large values domain can hardly be explained using a simple model.

The observed divergence could be due to imperfect model or may result from the dynamic processes in a language. The analysis of the data was performed for different time intervals to verify which of these possibilities is in play. The data selected using a 50 year sliding time window were fitted by power dependence. The dependence variations of Heaps' exponent with the time obtained for the English, Russian, German, and French languages are shown in figure 4. Two features of the obtained curves should be mentioned. First, a descending trend is typical of all the languages. Secondly, quasi-periodic variations with characteristic time periods of 60-100 years can be seen in the plots.

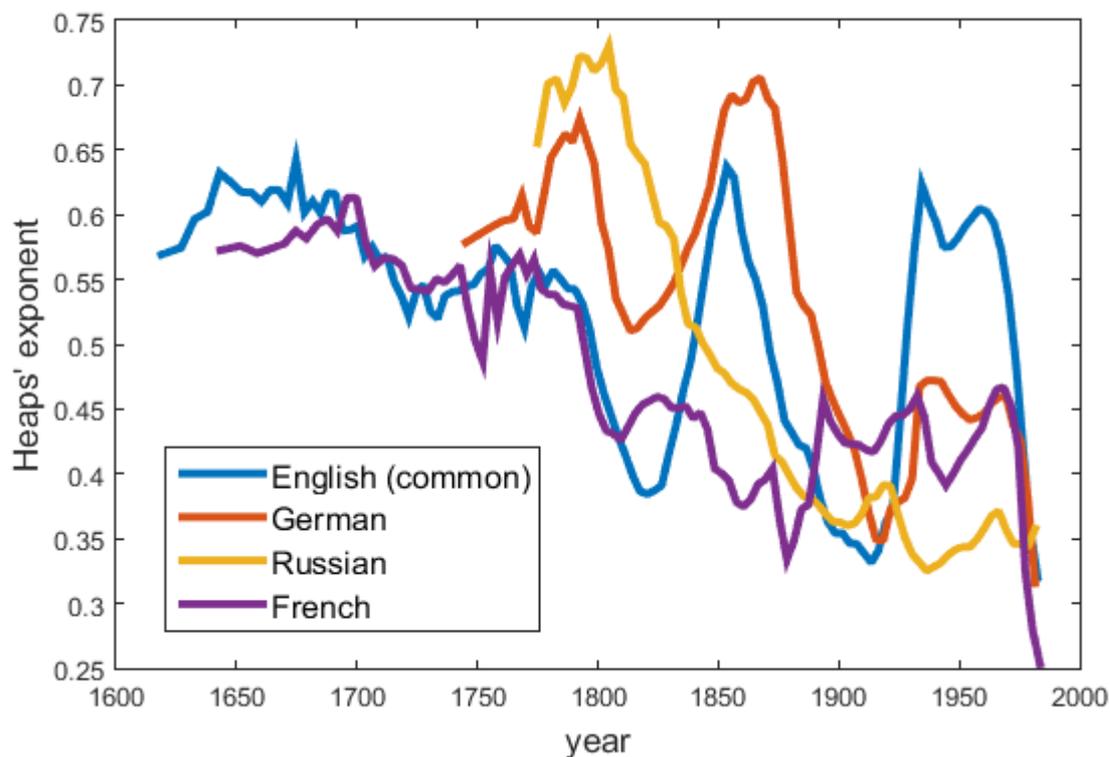

**Figure 4.** Change of the Heaps exponent with the time for the four European languages

To be sure that the power exponent variations which we can see in the figure are not erroneously obtained during the processing, we approximated the data by power dependence for the selected periods individually. Four years were selected which corresponded to the extreme points of the Heaps exponent time dependence, as well as 51-year intervals whose midpoints correspond to the selected years. The results are shown in Figure 5. Empirical values are shown by different markers which relate to different time intervals and their approximation by power dependence is shown by the lines. As can be seen from the figure, a change of the power exponent is unquestionable.

On average the number of texts in the Google Books Ngram base is much higher for later time periods. So the question is: can the observed variation of the Heaps exponent over time be due to the increase of the average size of the corpus or is it due to some dynamic processes in the language? The modelling results of the expected Heaps exponent at different text size are shown in figure 6. The modelling was performed using empirical word-usage frequencies and the modelled frequencies correlated with the Zipf law. The dependence $N(L)$ is approximated using the power law $N \sim L^{-k(L)}$ for each small section, where *k(L)* is assumed to be a slowly varying function. The required power exponent *k(L)* can be calculated by formula in (4).

$$k(L) = -\frac{\partial \ln N}{\partial \ln L} \qquad \qquad 4)$$

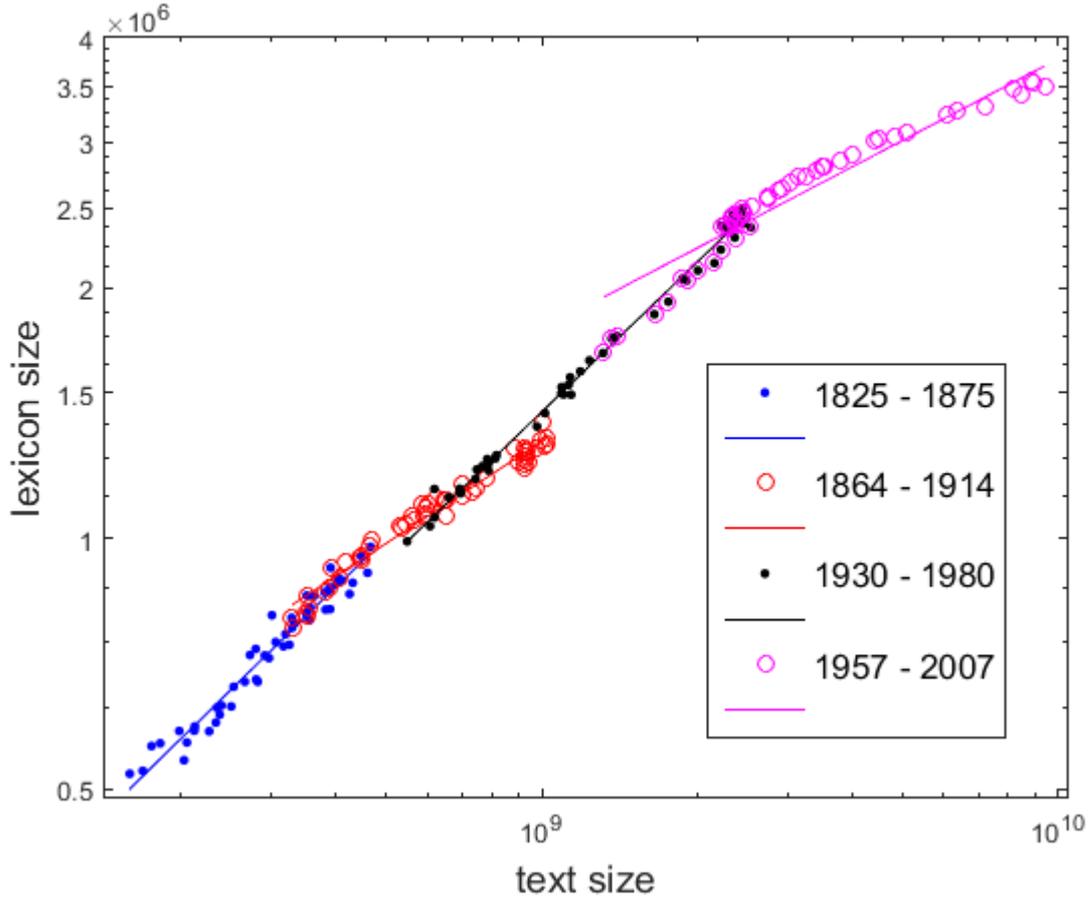

**Figure 5.** Dependence of the English language lexicon size on the text size
at different time periods

The Heaps exponents were modelled by (2,4) using empirical word-usage frequencies English for the year 2000 (the solid line). A descending trend and the Heaps exponent tending to zero at $L \to \infty$ can be observed in the plot. This is not surprising because our model has the finite potential lexicon and the dependence *N(L)* should tend to saturation. Two steps are seen in the plot. The red and yellow lines in the figure 6 represent the modelling results under the assumption that the word-usage frequencies corresponds to Zipf's law $p_k = Ak^{-\beta}$ (with the exponent β equal to respectively 1.077 and 1.698) with the finite number of $3.97 \cdot 10^6$ words. The curve tends asymptotically to zero as in the previous case. At that, almost a horizontal section of the local Heaps exponent can be seen in the plot close to the value $k(L) = 1/\beta$ as it was expected from the above-mentioned thoughts about the connection of the exponents of the Zipf and Heaps laws. Thus, it can be expected that two flat sections of the first curve correspond to the two segments in the word frequency plot which approach power laws (figure 1).

A comparison of Figure 4 and Figure 6 shows that the Heaps exponent decreases as the corpus size increases. But both model (2) and model (3) can (in contrast what is shown in Figure 4) only give monotonically decreasing dependencies for the Heaps exponent. Thus, the quasi-periodic variations of the Heaps exponent observed in figure 4 are most likely due to dynamic processes in the language.

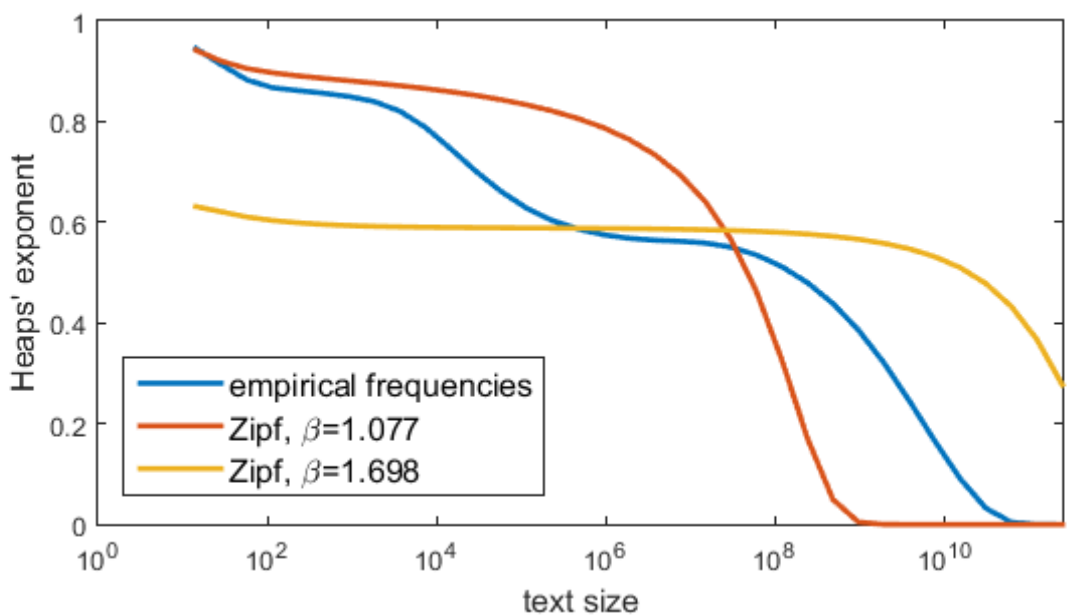

**Figure 6.** Modelled values of the Heaps exponent at different corpus size. The blue line represents the empirical frequency distribution (English, 2000), the red and yellow lines represent the Zipf distribution with different exponents

### 3. Conclusion
In summary we can observe that Heaps' law is fulfilled restrictively for small texts and texts pertaining to short historical periods. The phenomenon of $N(L)$ dependence saturation should be taken into account in case of large text corpora. Lexical dynamics should also be considered in case of diachronic text corpora. The analysis of the empirical data presented in the Google Books Ngram corpora shows that the Heaps exponent varies with characteristic time interval of 60-100 years, something which must reflect dynamic processes in the language.

This work was supported by the Russian Foundation for Basic Research (grant №12-06-00404a).